\documentclass[letterpaper,10pt,journal]{IEEEtran} 

\IEEEoverridecommandlockouts                              

\usepackage{graphicx}
\usepackage{amsmath}
\usepackage{amssymb}
\usepackage{booktabs}
\usepackage{tabularx}
\usepackage{array}
\usepackage{multirow}
\usepackage{xcolor}

\newcommand{\rev}[1]{#1}

\newcolumntype{Y}{>{\raggedright\arraybackslash}X}

\title{\LARGE \bf
Global-Local Attention Decomposition for Terrain Encoding in Humanoid Perceptive Locomotion
}

\author{Shengcheng Fu$^{1,2}$, Yang Zhang$^{3}$, Zhanxiang Cao$^{3,2}$, Liyun Yan$^{3,2}$, Yizhi Chen$^{1,2}$, Yunpeng Yin$^{4}$, Yue Gao$^{3,2*}$ \\
\vspace{0.15cm}
\normalsize $^{1}$Tongji University, Shanghai, China \\
\normalsize $^{2}$Shanghai Innovation Institute, Shanghai, China \\
\normalsize $^{3}$Shanghai Jiao Tong University, Shanghai, China \\
\normalsize $^{4}$Humanoid Robot (Shanghai) Co., Ltd., Shanghai, China \\
\normalsize $^{*}$Corresponding author: Yue Gao (email: yuegao@sjtu.edu.cn)
}

\begin{document}

\maketitle
\thispagestyle{empty}
\pagestyle{empty}


\begin{abstract}

Although reinforcement learning has significantly advanced humanoid locomotion, perceptive policies still struggle on sparse-foothold terrain and constrained environments. Success in these scenarios requires both broad terrain awareness and precise foothold selection, two perceptual roles that conventional encoders often entangle. To address this challenge, we propose Global-Local Attention Decomposition (GLAD) for terrain encoding in humanoid locomotion. Realized by a coarse-to-fine encoder over a robot-centric elevation map, GLAD explicitly separates these objectives: a global attention branch uses attention pooling to summarize the surrounding terrain context, while a local attention branch sparsifies the local features by terrain saliency and applies state-conditioned attention to encode precise foothold-relevant geometry. This explicit attention decomposition prevents the dilution of fine-grained spatial cues while reducing training overhead. Experiments demonstrate that GLAD enables reliable locomotion over challenging gaps, stepping stones, and stairs. Furthermore, the learned policy exhibits emergent terrain-responsive behaviors, autonomously following narrow paths and avoiding obstacles under forward-velocity commands alone, without explicit navigation planners. In real-world deployment on a Unitree G1 humanoid robot using onboard LiDAR, the proposed method achieves robust zero-shot sim-to-real transfer across diverse sparse-foothold and obstacle-rich domains.

\end{abstract}


\section{INTRODUCTION}

Humanoid robots, owing to their anthropomorphic body plan and legged mobility, are promising platforms for operating in environments designed for humans and adapting to diverse terrain conditions \cite{wu2024survey,tong2024advancements}. Reinforcement learning (RL) has recently driven substantial progress in humanoid locomotion, and even ``blind'' policies can now achieve robust walking over a wide range of ground conditions \cite{radosavovic2024real,radosavovic2024learning}. Nevertheless, reliable locomotion on complex terrain remains challenging because the robot must simultaneously perceive surrounding geometry, make terrain-aware decisions, and maintain robust control under limited stability margins \cite{duan2024learning,gadde2025no}. This challenge becomes especially pronounced on sparse-foothold terrain, where success depends on identifying feasible support regions and placing the feet precisely \cite{zhang2024learning}.

To handle such terrains, perceptive locomotion policies must incorporate exteroceptive observations such as elevation maps or depth images. While essential, these raw observations are high-dimensional and noisy, which can reduce robustness, increase computational cost, and complicate policy training \cite{long2025learning}. They therefore need to be encoded into compact terrain representations before being provided to the locomotion controller. However, conventional terrain encoders typically treat perception as an isolated geometric compression task, largely ignoring the robot's immediate kinematic state \cite{he2025attention}. Because this encoding process lacks state awareness, it tends to dilute fine-grained geometric cues, weaken the spatial correspondence between encoded features and reachable foothold regions, and ultimately limit foothold precision in RL-based perceptive locomotion.

\begin{figure}[!t]
\centering
\includegraphics[width=\columnwidth]{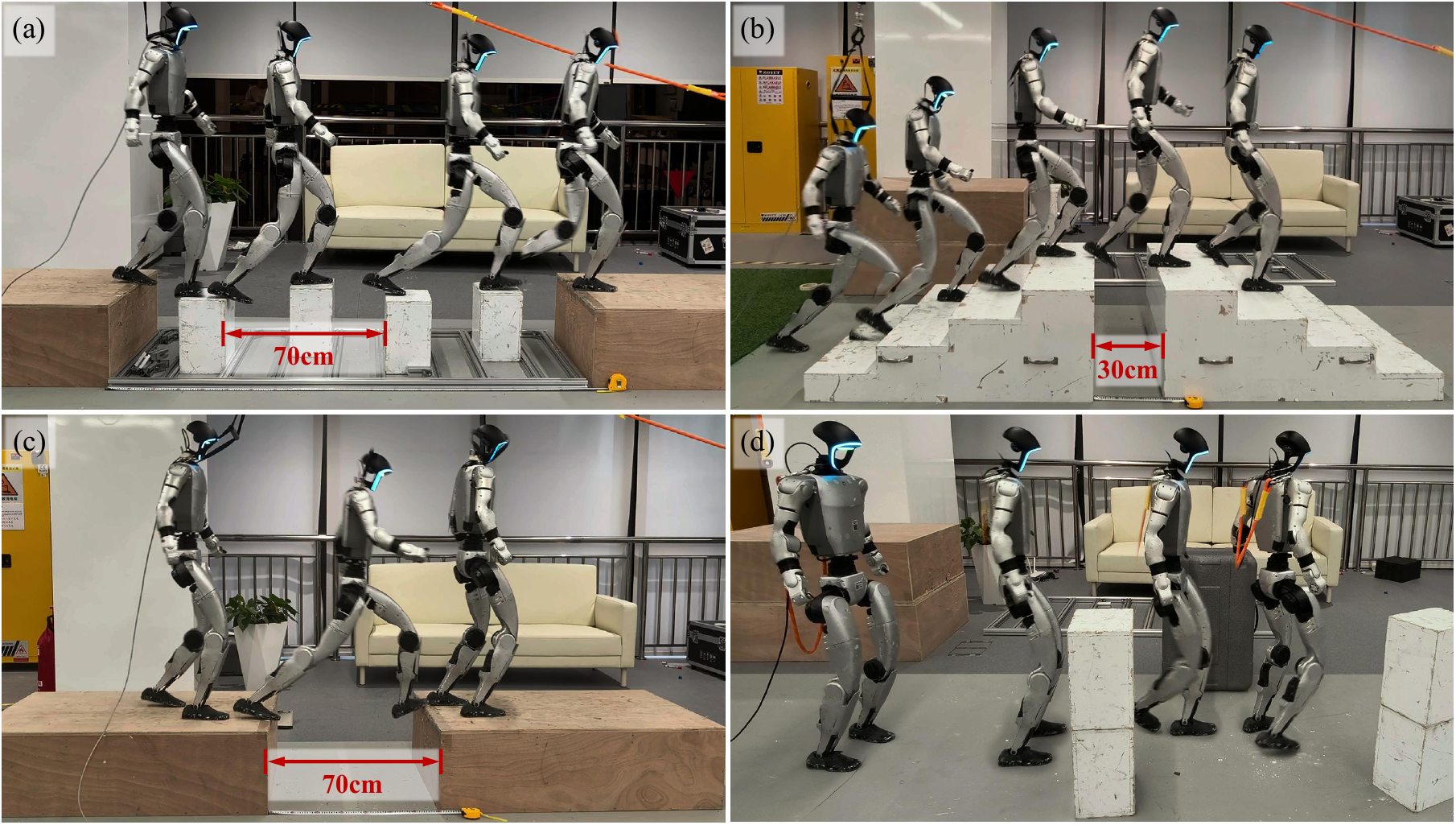}
\caption{Real-world locomotion results on the Unitree G1 humanoid robot. A single GLAD policy traverses sparse stepping stones, staircase--gap composite terrain, wide gaps up to $70$~cm, and dense obstacle-rich terrain.}
\label{fig:rel}
\end{figure}

This limitation is particularly problematic for sparse-foothold locomotion. To compensate for imprecise terrain representations, previous learning-based approaches often rely on terrain-specific curricula, expert policies, or handcrafted foothold rewards \cite{wang2025beamdojo,zhu2026hiking,zhang2026rpl}. Although effective, these strategies inject strong human priors and can reduce generalization across varied terrain layouts. Attention mechanisms offer a principled way to fuse proprioception with exteroception: guided by the current motion state, they can emphasize locomotion-relevant terrain regions and thereby support general, interpretable terrain encoding \cite{yang2021learning}. Building on this idea, Attention-based Map Encoding (AME) \cite{he2025attention} uses multi-head attention (MHA) to aggregate local terrain features conditioned on proprioception, enabling implicit and precise foothold selection; AME-2 \cite{zhang2026ame} further augments this design with global terrain context, letting the attended region adapt to the terrain type and improving zero-shot generalization to mixed and unseen terrain. However, these existing attention-based methods typically entangle multiple perceptual objectives within a single attention mechanism. They do not explicitly separate the two essential perceptual roles required for perceptive locomotion: capturing broad surrounding terrain context and encoding precise foothold-relevant local geometry.

To address this issue, we propose Global-Local Attention Decomposition (GLAD) for terrain encoding in humanoid locomotion. Realized by a coarse-to-fine encoder, GLAD explicitly separates the extraction of broad context from fine-grained foothold selection. A convolutional neural network (CNN) first extracts spatially aligned local features from a robot-centric elevation map. Subsequently, a global attention branch uses attention pooling to summarize the surrounding terrain context, while a local attention branch sparsifies these features by terrain saliency and applies state-conditioned MHA to encode precise, foothold-relevant local regions. Without explicit attention supervision, these two branches learn complementary behaviors: the global attention focuses on key terrain regions ahead of the robot, whereas the local attention highlights foothold-relevant regions. Because the surrounding context is aggregated by a separate branch, the fine-grained local geometry is no longer diluted.

Beyond sparse-foothold traversal, GLAD substantially enhances terrain-responsive locomotion in highly constrained environments. Guided solely by forward-velocity commands, the learned policy interprets terrain geometry to autonomously follow narrow paths and avoid nearby obstacles, exhibiting emergent local-navigation behaviors without requiring hierarchical planners. We validate GLAD in simulation and on a Unitree G1 humanoid robot, where it transfers zero-shot using only onboard LiDAR-based elevation mapping, as shown in Fig.~\ref{fig:rel}.

In summary, our main contributions are threefold:
\begin{itemize}
\item We propose GLAD, a coarse-to-fine terrain encoder for humanoid locomotion that explicitly decomposes attention into global terrain-context aggregation and state-conditioned local foothold-relevant encoding. This design enables the policy to jointly reason about broad traversability cues and precise support-region geometry from robot-centric elevation maps.
\item We show that assigning surrounding-terrain aggregation to the global branch allows the local attention to concentrate more strongly on foothold-relevant terrain features and remain closely aligned with the realized footholds, yielding complementary and interpretable attention behaviors. Saliency-based local feature sparsification further reduces the cost of dense attention over the full terrain feature set. The resulting policy improves sparse-foothold traversal and terrain-responsive locomotion, including emergent narrow-path following without hierarchical planning.
\item We validate GLAD in both simulation and real-world deployment on a Unitree G1 humanoid robot. Using only onboard LiDAR-based elevation mapping, without pre-mapping or external motion capture, the learned policy achieves robust zero-shot sim-to-real transfer across diverse sparse-foothold and obstacle-rich environments.
\end{itemize}

\begin{figure*}[!t]
\centering
\includegraphics[width=\textwidth]{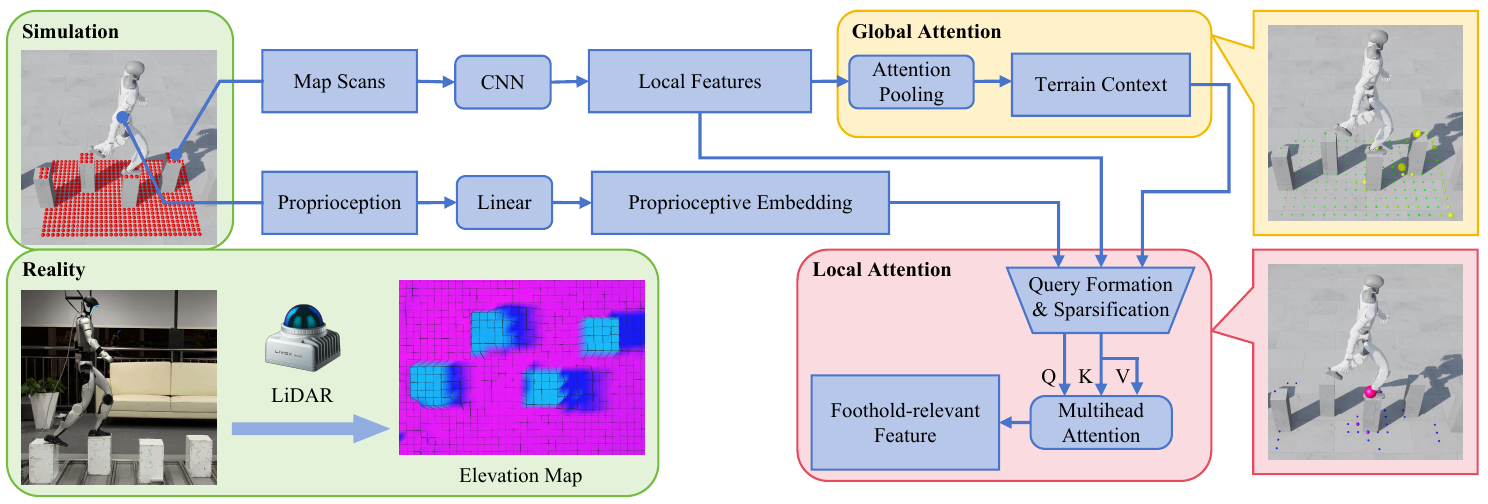}
\caption{Overview of the proposed Global-Local Attention Decomposition (GLAD) architecture. A CNN first extracts spatially aligned local features from a robot-centric elevation map. A global attention branch uses attention pooling to summarize the surrounding terrain context. This context is then fused with a proprioceptive embedding to form a state-conditioned query for MHA over saliency-sparsified local features, selectively encoding precise foothold-relevant geometry. The resulting explicitly decomposed representation is used by the locomotion policy for terrain-aware control.}
\label{fig:method_framework}
\end{figure*}

\section{RELATED WORK}

\subsection{Learning-based Humanoid Locomotion}

Recent advances in learning-based humanoid locomotion have been enabled in part by scalable training pipelines inherited from quadruped locomotion, which support massive simulation parallelization and reliable policy transfer to real robots \cite{rudin2022learning}. With only proprioceptive sensing, ``blind'' policies can already achieve robust walking over uneven ground, simple obstacles, and external disturbances \cite{radosavovic2024real,radosavovic2024learning}. For more challenging terrain, however, exteroceptive perception becomes increasingly important because the robot must anticipate hazards before contact and adapt its motion to surrounding geometry \cite{roychoudhury2023perception,de2024humanoid,luo2024pie}.

Existing perceptive locomotion methods mainly rely on depth images or elevation maps \cite{sun2025dpl}. Depth images are appealing for deployment, but in learning-based pipelines they often require costly rendering or additional temporal modeling to compensate for occlusion and limited field of view \cite{gadde2025no,zhu2026hiking,zhang2026rpl}. Elevation maps provide a more structured robot-centric description of local terrain geometry and can be generated efficiently in simulation or constructed on hardware from depth cameras or LiDAR \cite{duan2024learning,long2025learning,wang2025beamdojo,sun2025dpl,sun2025learning,song2025gait}. For sparse-foothold locomotion tasks, elevation maps are still often needed as privileged information during training or as an intermediate representation in the depth-image pipeline, because they provide precise terrain geometry that is critical for accurate foothold reasoning \cite{gadde2025no,zhang2026rpl,yu2024walking}.

\subsection{Locomotion on Sparse Footholds}

Discrete terrains with sparse footholds highlight the mobility advantage of legged robots, but they are also among the most challenging settings for perceptive locomotion, especially for humanoids. Model-based methods explicitly plan feasible foothold sequences within hierarchical perception--planning--control pipelines, achieving precise motion on sparse supports but remaining limited by model accuracy and robustness to perception and control errors \cite{jenelten2022tamols,grandia2023perceptive}. End-to-end RL has become a dominant direction due to its robustness and strong real-world performance, yet accurate foothold placement on complex terrain remains difficult to guarantee. A key bottleneck is terrain encoding: encoders based on multilayer perceptrons (MLPs), CNNs and latent-variable models directly compress elevation maps or depth images, often overlooking the terrain details and motion-state relevance required for foothold decisions. Therefore, prior learning-based methods for sparse footholds often rely on terrain-specific curricula, expert policies, or carefully designed foothold rewards \cite{zhang2024learning,yu2024walking,wang2025beamdojo,zhu2026hiking,zhang2026rpl}, which are difficult to engineer, inject strong human priors, and can bias policies toward conservative or terrain-specific foothold choices.

State-conditioned attention offers a promising alternative for implicit foothold planning. By fusing proprioception with exteroceptive terrain features, attention can focus on regions most relevant to the current locomotion decision and provide a more precise, general terrain representation \cite{yang2021learning}. AME \cite{he2025attention} applies MHA to local elevation-map features to obtain foothold-relevant focus without explicit foothold supervision. AME-2 \cite{zhang2026ame} observes that a query built from proprioception alone carries no prior about the surrounding terrain, and therefore adds an MLP with global max pooling that injects global terrain context into both the attention query and the encoder output. The attended region can then adapt to the terrain type rather than to the robot state alone, which improves zero-shot generalization to mixed and unseen terrain while preserving agile motion. However, both methods still apply a single MHA module to the full local feature set, forcing one attention mechanism to handle both global context aggregation and concentrated foothold-relevant encoding. This coupling increases computational cost and can weaken attention concentration and interpretability, motivating terrain representations that explicitly separate global context aggregation from state-conditioned local encoding.

\section{METHOD}

\subsection{Problem Formulation}

We formulate humanoid perceptive locomotion as a partially observable Markov decision process (POMDP)
\begin{equation}
\mathcal{M} = (\mathcal{S}, \mathcal{A}, \mathcal{O}, p, r, \gamma),
\end{equation}
where $\mathcal{S}$ is the state space, $\mathcal{A}$ is the action space, $\mathcal{O}$ is the observation space, $p(s_{t+1}\mid s_t,a_t)$ denotes the system dynamics, $r(s_t,a_t)$ is the reward function, and $\gamma\in(0,1)$ is the discount factor. The process is partially observable due to limited sensor field-of-view and measurement noise.

At time $t$, the environment is in a latent state $s_t\in\mathcal{S}$ that captures both the robot state and the surrounding terrain. The policy receives an observation $o_t\in\mathcal{O}$ composed of proprioceptive measurements and exteroceptive terrain perception (e.g., an elevation map), and outputs an action $a_t\in\mathcal{A}$. The policy is parameterized by $\theta$ and written as $a_t \sim \pi_{\theta}(\cdot \mid o_t)$.

The learning objective is to maximize the expected discounted return
\begin{equation}
J(\pi_{\theta}) = \mathbb{E}_{\pi_{\theta}}\left[\sum_{t=0}^{\infty} \gamma^{t} \, r(s_t,a_t)\right].
\end{equation}

\subsection{Observation Space}

The policy network observes proprioceptive information and elevation map scans in the robot-centric base frame. The actor observation includes the base angular velocity $\omega_t^{b}$, the gravity direction vector $g_t^{b}$, \rev{the velocity command $v_t^{\mathrm{cmd}}=[v_{x,t}^{\mathrm{cmd}},\,v_{y,t}^{\mathrm{cmd}},\,\omega_{z,t}^{\mathrm{cmd}}]$,} the joint positions $q_t$, the joint velocities $\dot{q}_t$, the previous action $a_{t-1}$, and an elevation map $m_t$ surrounding the robot base.

We denote the map scan as $m_t \in \mathbb{R}^{L\times W\times 3}$, where $L$ and $W$ are the numbers of map cells along the length and width directions, and each grid cell stores 3D coordinates expressed in the robot base frame.

The actor observation is written as
\begin{equation}
 o_t^{\mathrm{actor}} = \left\{\omega_t^{b},\, g_t^{b},\, \rev{v_t^{\mathrm{cmd}},\,} q_t,\, \dot{q}_t,\, a_{t-1},\, m_t\right\}.
\end{equation}

The critic observes the same information but without noise, and additionally receives the base linear velocity $v_t^{b}$:
\begin{equation}
 o_t^{\mathrm{critic}} = \left\{v_t^{b},\, \omega_t^{b},\, g_t^{b},\, \rev{v_t^{\mathrm{cmd}},\,} q_t,\, \dot{q}_t,\, a_{t-1},\, m_t\right\}.
\end{equation}

\subsection{Action Space}

The action $a_t\in\mathcal{A}$ is a 29-dimensional vector corresponding to the 29 actuated joints of the humanoid robot. For stable policy outputs, we interpret $a_t$ as an offset added to a nominal standing joint configuration $\theta_{\mathrm{stand}}$. The target joint position is thus $\theta_{\mathrm{target}} = \theta_{\mathrm{stand}} + a_t$.

\subsection{Global-Local Terrain Encoding}

As illustrated in Fig.~\ref{fig:method_framework}, the elevation-map input is first processed by the proposed GLAD encoder to form a compact terrain representation. This representation is then concatenated with the proprioceptive input and fed into the policy network.

\subsubsection{Spatially Aligned Local Feature Extraction}

Given the input elevation map $m_t \in \mathbb{R}^{L \times W \times 3}$, we first employ a two-layer CNN to extract spatially aligned local features. The 3D coordinates at each map location serve as the three input channels. Unlike AME \cite{he2025attention}, which convolves only the height values and then concatenates the original 3D coordinates with the CNN output to inject positional information, we directly convolve the 3D coordinates so that positional cues are embedded in the learned features from the beginning. 

Both convolutional layers use zero padding to preserve boundary information and maintain spatial alignment with the underlying terrain map. This spatial correspondence is important because the subsequent global and local attention modules should remain grounded to physical map locations. Without zero padding, the link between learned features and terrain locations becomes weaker, which negatively affects the final policy performance \cite{he2025attention}.

The first convolutional layer performs strided downsampling with a stride of $s$. Unlike AME \cite{he2025attention}, which preserves a one-to-one correspondence between CNN features and the original scan points, our encoder decouples terrain sensing resolution from encoded feature resolution. The resulting features remain aligned with the original elevation map, but on a coarser grid of size $\lceil L/s \rceil \times \lceil W/s \rceil$. This design shortens the local feature sequence, reduces the computational cost of the subsequent attention modules, and allows higher-resolution elevation maps to be used as input while maintaining a similar encoded feature resolution for downstream processing.

The second convolutional layer projects the downsampled features into a unified embedding dimension $D$, which is shared by the proprioceptive embedding, the surrounding terrain context vector, and the MHA components. The CNN thus outputs a feature grid $F_{\text{grid}} \in \mathbb{R}^{\lceil L/s \rceil \times \lceil W/s \rceil \times D}$. These local features encode both terrain geometry and positional information around each retained map location. This is important for foothold selection, since a feasible foothold depends not only on local terrain geometry but also on whether the corresponding location is reachable by the robot.

\subsubsection{Global-Local Attention Decomposition}

After local feature extraction, the GLAD terrain encoding is constructed in three steps: attention pooling aggregates all local features into a compact surrounding terrain context vector; a saliency score sparsifies the local features to a retained-feature budget of $K$; and the context is fused with a proprioceptive embedding to form a state-conditioned query that drives MHA over the retained features.

Let $\{k_i\}_{i=1}^{N}$ denote the flattened local terrain features extracted by the CNN, where $N = \lceil L/s \rceil \times \lceil W/s \rceil$ is the sequence length and $k_i\in\mathbb{R}^{D}$. To summarize the surrounding terrain information, we apply a linear layer to each local feature to produce a global-attention logit
\begin{equation}
u_i = v^{\top} k_i + b_u,
\end{equation}
where $v$ and $b_u$ are learnable parameters. The normalized global-attention weights are then computed as
\begin{equation}
\beta_i = \frac{\exp(u_i)}{\sum_{j=1}^{N}\exp(u_j)},
\end{equation}
and the surrounding terrain context feature $c \in \mathbb{R}^{D}$ is obtained by a weighted sum:
\begin{equation}
c = \sum_{i=1}^{N} \beta_i k_i.
\end{equation}
This global branch summarizes the local terrain features into a compact representation of the surrounding terrain context.

Because the local features already encode each cell's position in the robot base frame, saliency can be assessed from the features themselves. The local branch therefore ranks the features by a learned saliency score
\begin{equation}
s_i = w^{\top} k_i + b_s, \quad i=1,\ldots,N,
\end{equation}
where $w$ and $b_s$ are learnable parameters. Whereas the global logits $u_i$ are normalized into the weights forming $c$, the scores $s_i$ act only through their ordering: we retain the top-$K$ local features and discard less salient regions. During training, Gumbel perturbations are applied to the scores, and straight-through softmax weighting ($\tau=1$) propagates gradients through the discrete selection; inference uses deterministic hard top-$K$ selection. This coarse prefilter removes clearly irrelevant regions before the fine-grained attention stage. We then concatenate $c$ with the $D$-dimensional proprioceptive embedding and project the result into a state-conditioned query vector $q\in\mathbb{R}^{D}$ via a linear layer. The retained features are processed by MHA as key and value, using $q$ as the query, to yield the fine-grained, foothold-relevant feature $f \in \mathbb{R}^{D}$.

Finally, the surrounding terrain context $c$ and the foothold-relevant feature $f$ are concatenated to form the unified terrain representation fed into the locomotion policy. In this way, global attention summarizes the surrounding terrain context, while local attention performs state-conditioned encoding over a sparse set of salient local regions. This explicit separation avoids the cost of applying MHA to the full terrain feature set, lowers training overhead, and improves the effectiveness and interpretability of the terrain representation for foothold-related decision making.

\section{EXPERIMENTS}

\subsection{Training and Experimental Setup}

\subsubsection{Baselines and Ablations}
We compare GLAD with two terrain-encoder baselines and two ablations. AME \cite{he2025attention}, the primary attention-based baseline, applies state-conditioned MHA to CNN-extracted local terrain features.

The second baseline, denoted as \textit{AME with Global Context} (AME-GC), adopts the global-context terrain-encoder design of AME-2 \cite{zhang2026ame}. Specifically, an additional MLP followed by max pooling produces a global feature, which is concatenated with both the MHA query and the encoder output. AME-GC adapts only the relevant encoder design rather than reproducing the complete AME-2 system or its original training pipeline.

Two ablations isolate the GLAD branches: \textit{GLAD w/o global attention} removes attention pooling but retains saliency-based sparsification and MHA, whereas \textit{GLAD w/o local attention} removes sparsification and MHA and retains only attention pooling.

\rev{All methods use identical training settings, and corresponding components across methods are matched in architecture and hyperparameters; only their terrain-feature aggregation designs differ.}

\subsubsection{Network Hyperparameters}
The actor and critic share the terrain encoder but use separate ELU-activated MLPs with hidden dimensions $[512,256,128]$.

The terrain encoder receives a $1.6\,\mathrm{m}\times1.0\,\mathrm{m}$ robot-centric elevation map at $0.05$~m resolution ($L=33$, $W=21$). Its two-layer CNN comprises a $3{\rightarrow}16$ convolution (kernel $5$, stride $2$, padding $2$) and a $16{\rightarrow}64$ convolution (kernel $3$, stride $1$, padding $1$), each followed by ReLU and batch normalization. The output is a $17\times11$ grid with embedding dimension $D=64$ and flattened length $N=187$.

Separate linear layers, without activation or normalization, project the $96$-dimensional actor and $99$-dimensional critic proprioceptive inputs to $D=64$. The single-layer MHA uses $16$ heads, the standard output projection, and no dropout, additional layer normalization, or residual connections.

AME, AME-GC, and GLAD use the same input definitions and identical settings for corresponding components. GLAD uses a retained-feature budget of $K=32$, while AME-GC adds an ELU-activated $[256,128]$ MLP followed by max pooling.

\subsubsection{Training Details}
All policies are trained in NVIDIA Isaac Sim using proximal policy optimization (PPO) \cite{schulman2017proximal} and a two-stage procedure following AME-based practice \cite{he2025attention}.

Stage 1 uses perfect terrain perception to initialize terrain encoding and basic locomotion. Its basic terrain set consists of stairs, random grids, random rough terrain, slopes, densely tiled stepping stones, and gaps.

\begin{figure}[!t]
\centering
\includegraphics[width=\columnwidth]{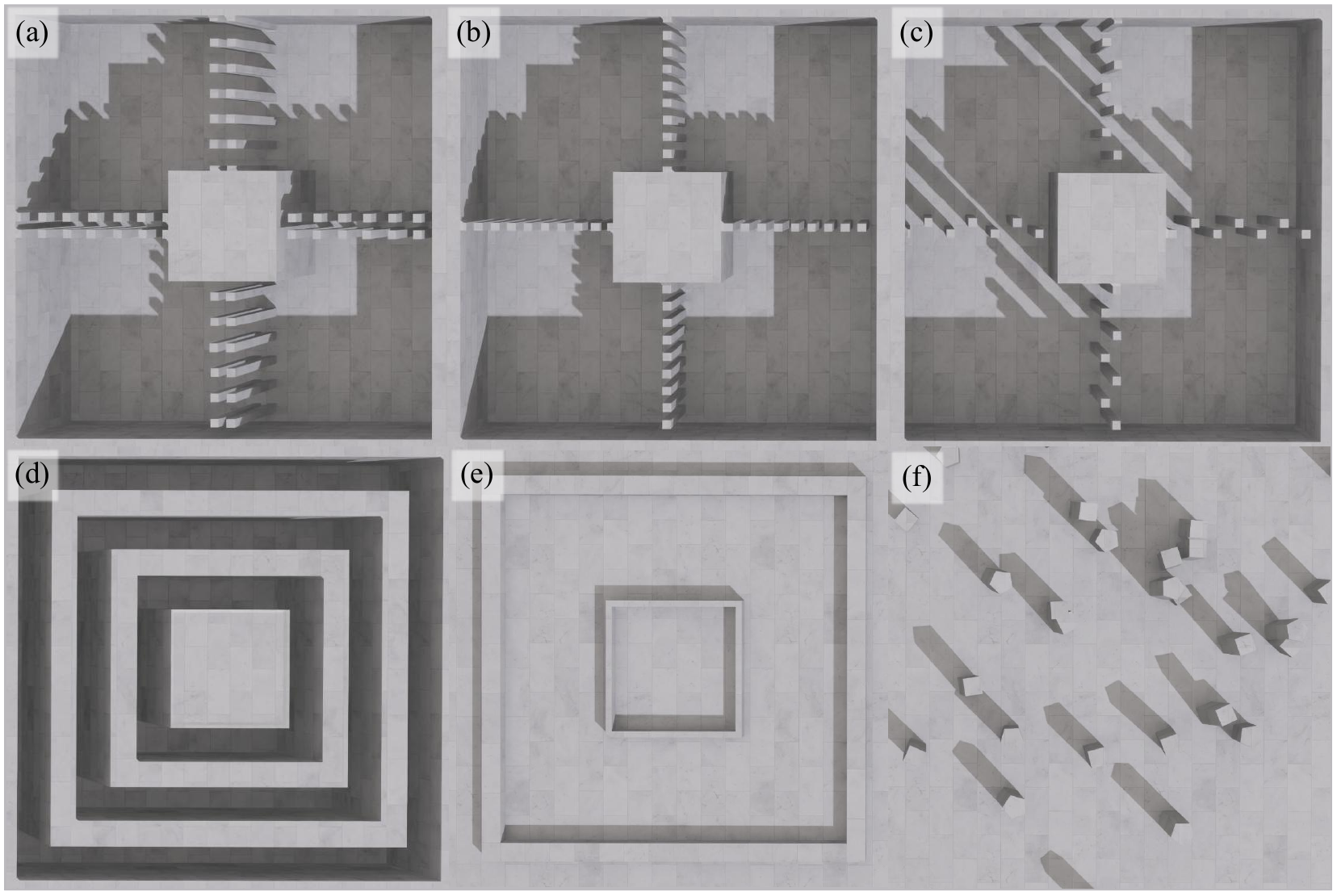}
\caption{Representative Stage-2 training terrains at the highest curriculum difficulty level. The terrain set includes sparse stepping stones, rails, gaps, and randomly distributed cylindrical obstacles.}
\label{fig:training_terrain}
\end{figure}

Stage 2 retains stairs and gaps but replaces densely tiled stepping stones with the harder parallel-row, single-row, and alternating stepping-stone layouts, and additionally introduces rails and randomly distributed cylindrical obstacles in Fig.~\ref{fig:training_terrain}. Observation noise and domain randomization are introduced mainly in this stage.

Both stages use a ten-level curriculum that promotes successful robots and demotes failed ones. In both stages, velocity commands are resampled every $10$~s, with $v_x^{\mathrm{cmd}}\sim\mathcal{U}(0,1.5)$~m/s and $v_y^{\mathrm{cmd}}=0$. The yaw-rate command $\omega_z^{\mathrm{cmd}}$ is not sampled directly but generated by a heading controller from the difference between a resampled target heading and the current base heading, and is clipped to $[-1,1]$~rad/s. Each method is trained with $4096$ parallel robots for $10{,}000$ learning iterations per stage.

We use standard locomotion rewards without handcrafted foothold rewards or terrain-specific shaping. Generic randomization includes per-step observation noise, terrain-map scan drift, external pushes, torso-mass variation, and friction variation, with no encoder-specific tuning. This isolates the contribution of general terrain representations from terrain-specific priors.

All training settings, including the PPO hyperparameters, are identical for GLAD, AME, AME-GC, and both ablations. All methods also share the same fixed random seed of $42$, which controls network initialization, procedural terrain generation, simulator randomization, and PPO sampling.

\subsection{Training Efficiency and Convergence}

\begin{figure}[!t]
\centering
\includegraphics[width=\columnwidth]{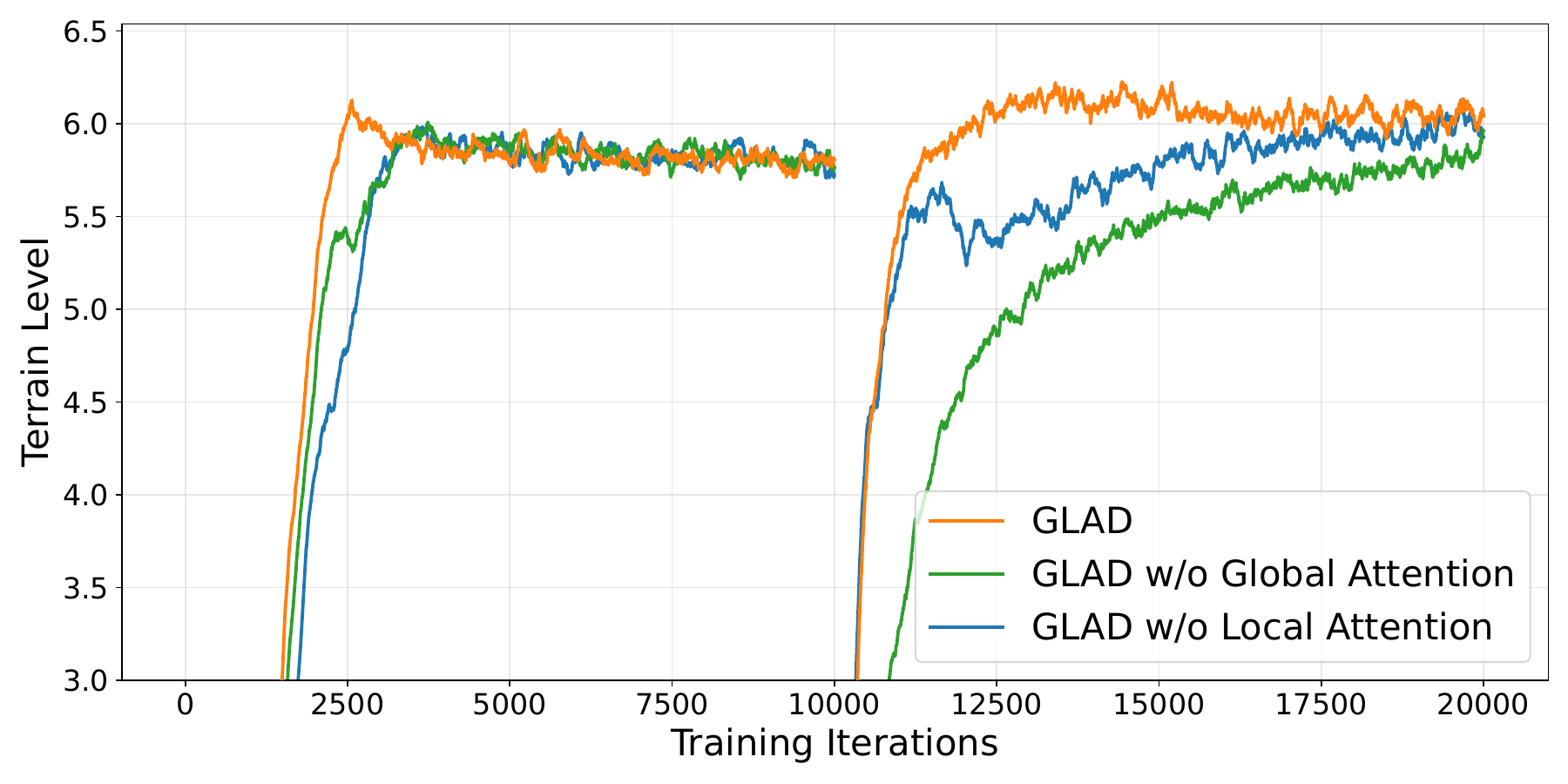}
\caption{Terrain-level curriculum progression during the two-stage training process for the proposed method and the two ablation variants. The full model reaches higher terrain levels earlier and exhibits the best convergence behavior.}
\label{fig:terrain_level}
\end{figure}

We first compare training performance through the terrain-level curriculum progression shown in Fig.~\ref{fig:terrain_level}. The full model consistently reaches higher terrain levels earlier than both \textit{GLAD w/o global attention} and \textit{GLAD w/o local attention}, indicating that removing either branch slows convergence. This trend suggests that global and local attention not only improve final performance, but also jointly accelerate curriculum progression and improve convergence.

We evaluate the computational efficiency during training on a single NVIDIA H200 GPU. Under the shared CNN strided-downsampling configuration described above, the proposed GLAD method requires only $1.00$ day to complete the two-stage training, compared with $1.36$ days for AME and $1.50$ days for AME-GC. This reduction emphasizes the efficiency advantage of our coarse-to-fine design, which avoids applying MHA to the full local feature set while retaining strong locomotion performance.

\rev{We also analyze training time across retained-feature budgets $K$. Relative to the $1.00$ day of the default $K=32$, raising $K$ to $64$ and $128$ increases it to about $1.12$ and $1.24$ days, whereas $K=16$ saves little, at $0.98$ days. As quantified in Section~\ref{subsec:attention_alignment}, the learned local attention concentrates on only about four locations, so moderate budgets already provide ample capacity, and varying $K$ at inference has negligible influence on task performance over $K=16$--$128$, degrading only for much smaller budgets. Capacity alone would therefore permit a smaller $K$. Early in training, however, the selection scores remain close to their random initialization, so the retained features form a nearly random subset and a smaller $K$ is correspondingly less likely to include a foothold-relevant location; this increased the learning difficulty and could prevent a stable foothold-aligned attention pattern from emerging. Since reducing $K$ to $16$ yields only a marginal efficiency gain, we set $K=32$ as a conservative default balancing candidate capacity, learning stability, and computational efficiency.}

\subsection{Simulation Experiments}

\subsubsection{Unseen Terrain Performance}

\begin{figure}[!t]
\centering
\includegraphics[width=\columnwidth]{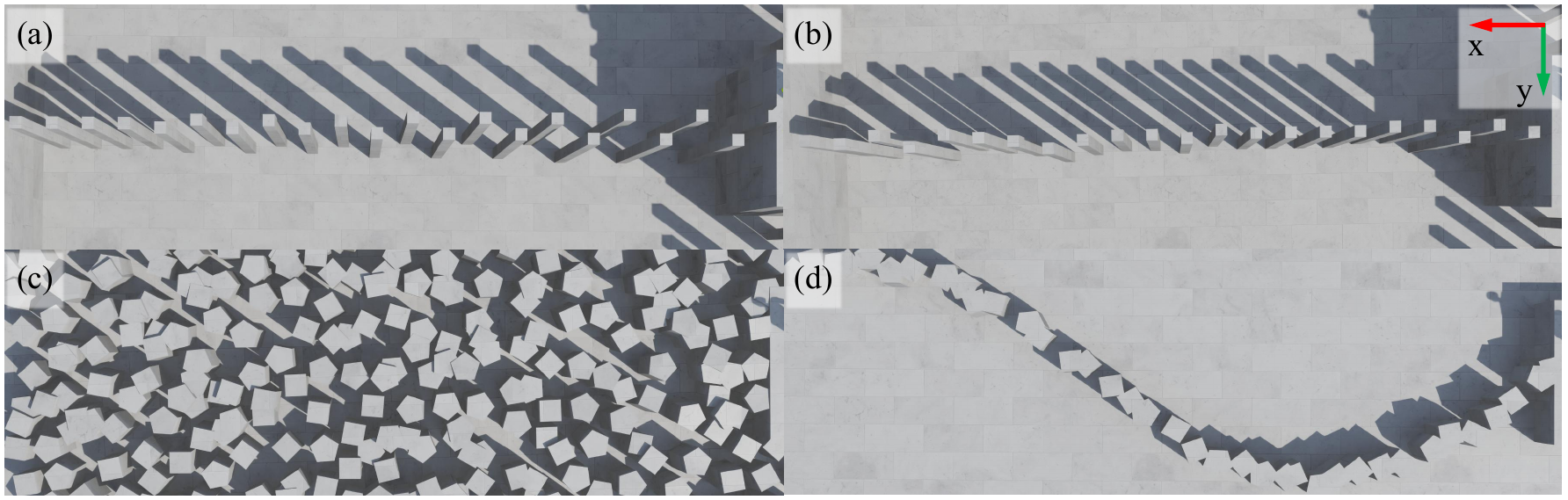}
\caption{Illustration of the four test terrains used in simulation: a composite stepping-stone course, a random single-row stepping-stone course, irregular pentagonal pillars, and a sinusoidal narrow path. These terrains are designed to evaluate both precise foothold placement and terrain-responsive locomotion under different spatial constraints.}
\label{fig:test_terrain}
\end{figure}

\begin{figure*}[!t]
\centering
\includegraphics[width=\textwidth]{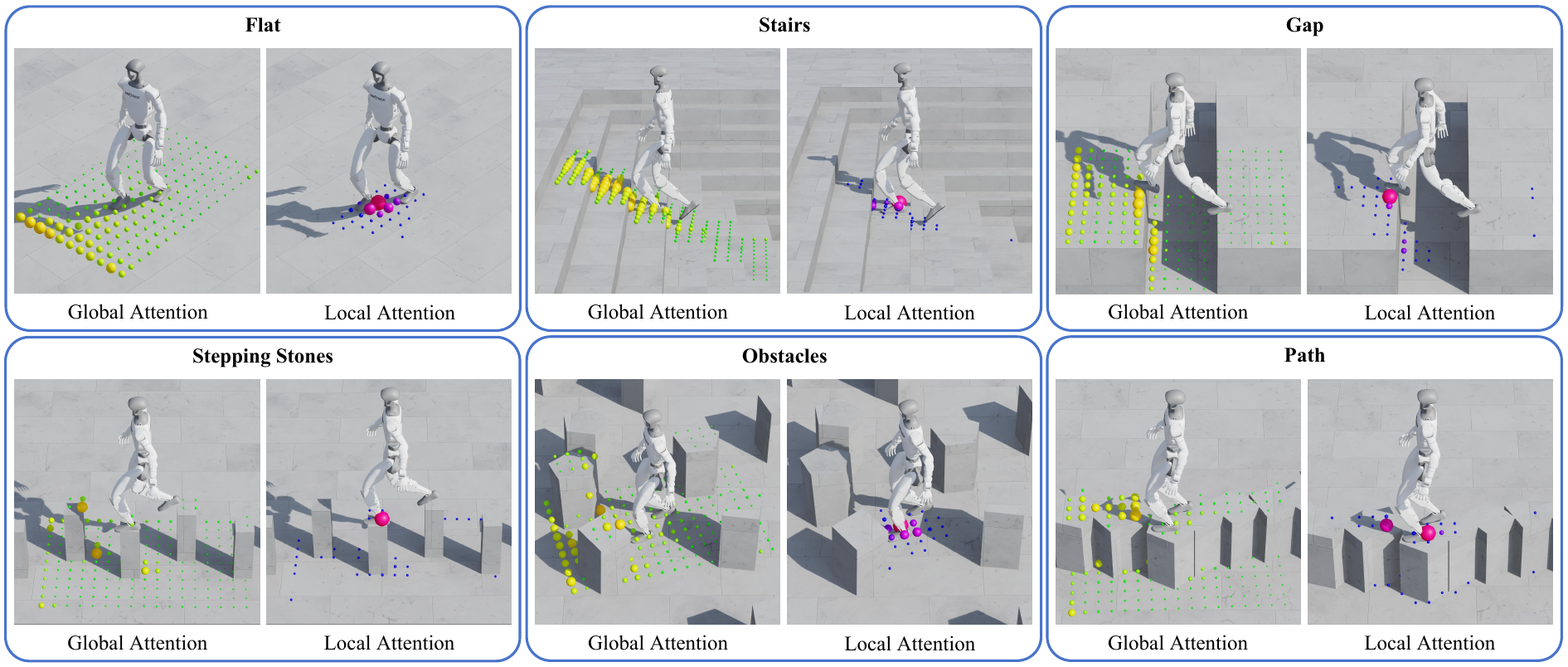}
\caption{Visualization of global and local attention weights of the proposed GLAD encoder. The displayed scan locations correspond to the spatially aligned local features, where yellow and red markers with higher intensity and larger size indicate larger global-attention and local-attention weights, respectively.}
\label{fig:attention_visual}
\end{figure*}

\begin{table}[!t]
\caption{Mean success rates (\%) and standard deviations over five evaluation seeds on the four simulation test terrains.}
\label{tab:sim_success_rate}
\centering
\scriptsize
\setlength{\tabcolsep}{1.2pt}
\begin{tabularx}{\columnwidth}{@{}>{\raggedright\arraybackslash}p{2.85cm}*{4}{>{\centering\arraybackslash}X}@{}}
\toprule
\multirow{2}{*}{Method} & \multicolumn{4}{c}{Test Terrain} \\
\cmidrule(lr){2-5}
 & a & b & c & d \\
\midrule
AME & 93.99{\tiny$\pm$2.04} & 94.77{\tiny$\pm$1.23} & 81.72{\tiny$\pm$3.22} & 53.67{\tiny$\pm$3.98} \\
AME-GC & 95.90{\tiny$\pm$0.62} & 92.58{\tiny$\pm$1.45} & 70.23{\tiny$\pm$4.20} & 81.13{\tiny$\pm$3.70} \\
GLAD w/o global attention & 81.99{\tiny$\pm$2.89} & 84.77{\tiny$\pm$1.49} & 87.70{\tiny$\pm$2.37} & 30.66{\tiny$\pm$4.40} \\
GLAD w/o local attention & 93.36{\tiny$\pm$0.57} & 94.53{\tiny$\pm$1.77} & 71.95{\tiny$\pm$5.03} & 49.69{\tiny$\pm$3.67} \\
GLAD & \textbf{96.64}{\tiny$\pm$0.80} & \textbf{97.03}{\tiny$\pm$1.18} & \textbf{92.42}{\tiny$\pm$1.52} & \textbf{93.48}{\tiny$\pm$1.98} \\
\bottomrule
\end{tabularx}
\end{table}

\rev{Since all methods solve the training terrains reliably, we construct four more challenging long-horizon test terrains with unseen instances, each forming a $10$~m course as shown in Fig.~\ref{fig:test_terrain}. Terrain (a) is a composite course that transitions from alternating double-row stepping stones to a single row, whereas Terrain (b) consists of a single row with random lateral offsets. The stepping stones in both terrains have $20~\mathrm{cm}\times20~\mathrm{cm}$ top surfaces, providing sparse yet clearly defined support regions that primarily test precise foot placement. Terrain (c) comprises randomly distributed pentagonal pillars, whose cluttered arrangement makes feasible footholds more difficult to identify. Terrain (d) places pentagonal pillars along a sinusoidal path, requiring the robot to maintain precise footholds within a constrained support region while making substantial directional adjustments. For all four terrains, all robots are initialized facing the positive $x$ direction and receive a forward-velocity command only, with $v_y^{\mathrm{cmd}}=0$ and $\omega_z^{\mathrm{cmd}}=0$. The commanded forward velocity is $1.0$~m/s for Terrains (a) and (b) and $0.8$~m/s for Terrains (c) and (d). Evaluation is repeated under five random seeds; for each terrain type and seed we generate $64$ distinct terrain instances and test $512$ robots in parallel. All methods use the same seeds, ensuring identical instances and matched simulator randomization. Table~\ref{tab:sim_success_rate} reports the mean success rate and standard deviation.}

On stepping stones such as Terrains (a) and (b), most compared methods achieve mean success rates above \(90\%\). However, \textit{GLAD w/o global attention} falls below $90\%$, confirming that relying solely on fine-grained foothold-relevant features is insufficient. The loss of surrounding terrain context degrades performance even on terrain layouts that are structurally similar to the training set.

A larger performance gap emerges on Terrains (c) and (d). In Terrain (c), feasible support regions are more abundant, but their cluttered arrangement makes foothold planning and selection more challenging. The full GLAD model combines foothold-level precision with surrounding terrain context to maintain a success rate above \(90\%\). By contrast, AME-GC performs noticeably worse, which may reflect the challenge of using a single MHA pathway to balance surrounding terrain context and precise foothold-relevant geometry. Similarly, \textit{GLAD w/o local attention} suffers a substantial performance drop, indicating that attention pooling alone cannot provide the fine-grained terrain representation required for precise foot placement in highly cluttered environments. Notably, \textit{GLAD w/o global attention} ranks second on this terrain, ahead of both baselines. Because feasible supports are densely and randomly distributed along the whole course, a nearby support is almost always available and the robot seldom needs to anticipate where the traversable region continues, so surrounding terrain context is least valuable here. This variant nevertheless remains clearly below the full model, and it collapses on Terrain (d), where the traversable region follows a curved path and sustained anticipation becomes essential.

Terrain (d) represents the more demanding setting, requiring anticipatory directional adjustments prior to reaching infeasible regions. Under this scenario, GLAD continues to outperform the other methods. Although AME-GC performs better than AME, demonstrating the benefit of the introduced global terrain context, its remaining gap to GLAD suggests that simply fusing this information into the same attention pathway is insufficient. The ablation results show that removing the local attention branch fails to yield sharp foothold cues, while removing the global attention branch strips away the contextual forward traversability needed for sustained path following. These findings confirm that the surrounding terrain context feature $c$ and the foothold-relevant feature $f$ fulfill complementary roles. By explicitly decomposing them, GLAD avoids architectural entanglement and leads to more robust, terrain-responsive legged locomotion.

\subsubsection{Attention--Foothold Alignment}
\label{subsec:attention_alignment}
To quantify whether local attention is both concentrated and aligned with actual footholds, we record the MHA attention weights at touchdown events and compare each attended local feature position $p_i$ with the realized foothold position $p_f$. We report the following metrics:
\begin{equation}
\begin{aligned}
D_{\mathrm{attn}} &= \sum_i \alpha_i \|p_i-p_f\|,\\
D_{\mathrm{peak}} &= \|p_{\arg\max_i \alpha_i}-p_f\|,\\
A_{\mathrm{peak}} &= \max_i \alpha_i,\\
H &= -\sum_i \alpha_i\log\alpha_i,\\
N_{\mathrm{eq}} &= \exp(H),
\end{aligned}
\end{equation}
where $\alpha_i$ is the MHA attention weight assigned to local feature $i$, averaged across the $16$ attention heads. $D_{\mathrm{attn}}$ is the attention-weighted foothold distance, $D_{\mathrm{peak}}$ and $A_{\mathrm{peak}}$ are the distance and weight of the most attended feature, and $H$ is the attention entropy. $N_{\mathrm{eq}}$ is the entropy-equivalent number of attended locations, i.e., the number of uniformly weighted locations that yields the same entropy; lower values indicate stronger concentration. Table~\ref{tab:attention_foothold_alignment} reports the mean and standard deviation of each metric over all touchdown events collected across the four terrain types in Fig.~\ref{fig:test_terrain}. Distances are in meters. All GLAD results use one policy trained with $K=32$, with $K$ varied only at inference.

\begin{table}[!t]
\caption{Attention--foothold metrics over all touchdown events across the four simulation test terrain types.}
\label{tab:attention_foothold_alignment}
\centering
\scriptsize
\setlength{\tabcolsep}{1.2pt}
\begin{tabularx}{\columnwidth}{@{}>{\raggedright\arraybackslash}p{2.1cm}*{5}{>{\centering\arraybackslash}X}@{}}
\toprule
Method & $D_{\mathrm{attn}}\downarrow$ & $D_{\mathrm{peak}}\downarrow$ & $A_{\mathrm{peak}}\uparrow$ & $H\downarrow$ & $N_{\mathrm{eq}}\downarrow$ \\
\midrule
GLAD ($K=16$)  & 0.120{\tiny$\pm$0.057} & 0.103{\tiny$\pm$0.057} & 0.539{\tiny$\pm$0.163} & 1.248{\tiny$\pm$0.416} & 3.77{\tiny$\pm$1.40} \\
GLAD ($K=32$)  & 0.118{\tiny$\pm$0.055} & 0.104{\tiny$\pm$0.059} & 0.537{\tiny$\pm$0.163} & 1.259{\tiny$\pm$0.421} & 3.82{\tiny$\pm$1.42} \\
GLAD ($K=187$) & 0.120{\tiny$\pm$0.059} & 0.104{\tiny$\pm$0.062} & 0.534{\tiny$\pm$0.166} & 1.266{\tiny$\pm$0.436} & 3.87{\tiny$\pm$1.55} \\
\midrule
AME    & 0.406{\tiny$\pm$0.082} & 0.100{\tiny$\pm$0.110} & 0.088{\tiny$\pm$0.037} & 4.581{\tiny$\pm$0.155} & 98.76{\tiny$\pm$14.89} \\
AME-GC & 0.514{\tiny$\pm$0.034} & 0.225{\tiny$\pm$0.268} & 0.045{\tiny$\pm$0.026} & 5.018{\tiny$\pm$0.122} & 152.06{\tiny$\pm$15.87} \\
\bottomrule
\end{tabularx}
\end{table}

For the reported settings from \(K=16\) to \(K=187\), GLAD maintains $N_{\mathrm{eq}}\approx4$, even when all 187 features are retained, showing that its concentration is learned rather than imposed by Top-$K$ truncation. For a feature-count-matched comparison, GLAD ($K=187$), AME, and AME-GC all attend over the same 187 features. Under this condition, GLAD yields substantially smaller $H$ and $N_{\mathrm{eq}}$, indicating a markedly more concentrated distribution, together with a smaller attention-weighted foothold distance; it therefore concentrates its attention near the realized foothold rather than merely producing a sparse distribution. AME attains the smallest $D_{\mathrm{peak}}$, so its most attended feature can also lie near the foothold; its much larger $D_{\mathrm{attn}}$ and $N_{\mathrm{eq}}$, however, show that the remaining attention mass is spread over distant locations. By separating environment aggregation from foothold encoding, GLAD lets its MHA provide a clearer foothold cue. AME instead balances both roles in one MHA pathway, while AME-GC injects additional global context into the same pathway and exhibits an even more diffuse attention distribution.

\subsubsection{Interpretable Attention Decomposition}

To better understand the attention mechanism, we visualize the global and local attention weights across various terrain layouts in Fig.~\ref{fig:attention_visual}. The resulting distributions exhibit a distinct functional separation between the two branches.

The global attention produces terrain-dependent, broadly receptive patterns. On flat ground, the weights lean toward the far end of the observation map. On stairs, gaps, and obstacle-rich environments, the attention maintains distant awareness while increasingly highlighting geometric edges with significant height variations. On more constrained terrains, such as stepping stones and narrow paths, it concentrates on nearby forward-traversable regions. This adaptability demonstrates that the surrounding terrain context feature $c$ effectively captures anticipatory cues regarding upcoming geometric changes and forward traversability.

Conversely, the local attention yields highly concentrated spatial patterns. Guided by the learned saliency scores, the retained local features concentrate primarily on near-field feasible foothold regions. The subsequent MHA weights typically peak near the robot's current or next foothold locations. This highly localized behavior confirms that the fine-grained, foothold-relevant feature $f$ directly provides the spatial geometry required for precise foot placement.

\subsection{Real-world Deployment}

We deploy the proposed method on a Unitree G1 humanoid robot. The robot weighs $35$~kg, stands $1.32$~m tall, and has $29$ actuated degrees of freedom. For exteroceptive perception, it is equipped with a Livox Mid-360 LiDAR that provides point clouds together with synchronized inertial measurement unit (IMU) measurements.

For deployment, we use FAST-LIO \cite{xu2021fast} to fuse LiDAR and IMU measurements for state estimation, and construct a robot-centric gridded elevation map from the fused point cloud \cite{miki2022elevation}. \rev{Invalid cells do not occur in simulation, while real-world map holes caused by LiDAR occlusion are filled using neighboring valid elevations before policy inference.} The system runs on a laptop equipped with an RTX 5090 Laptop GPU, where both the elevation-mapping module and the control policy operate at $50$~Hz.

As shown in Fig.~\ref{fig:rel}, we test four representative real-world terrains: alternating stepping stones with top surfaces measuring $20~\mathrm{cm}\times25~\mathrm{cm}$ and $70$~cm spacing between same-side stones; a composite staircase with $30$~cm treads, $12$~cm risers, and a $30$~cm gap between its ascending and descending segments; a $70$~cm wide gap, exceeding the maximum training gap of $50$~cm and approaching the forward sensing limit; and an obstacle-rich route containing wooden blocks and suitcases.

\rev{Each method is tested five times per terrain under the same protocol; success requires course completion without falling or human intervention. GLAD completes $19$ of $20$ trials, succeeding in all stair, gap, and obstacle trials and in four of five on the alternating stepping stones; AME-GC and AME reach $19$ and $18$ of $20$ under the same protocol. All failures occur on the alternating stepping stones, where LiDAR mixed-pixel artifacts at depth discontinuities blur elevation-map boundaries and can make adjacent stones appear connected, commonly limiting success across methods. Because the available physical courses are shorter and less constrained than the long-horizon simulation tests, they exhibit a ceiling effect that limits their ability to distinguish the terrain encoders. These tests therefore primarily validate zero-shot transfer and deployability, while the long-horizon simulation tests provide the main encoder comparison.}

\section{CONCLUSIONS}

We presented GLAD, a coarse-to-fine terrain encoder that separates global context aggregation from state-conditioned local foothold encoding. This decomposition reduces attention entanglement and computational cost while producing interpretable, foothold-aligned attention. Simulation demonstrated robust locomotion on challenging long-horizon test terrains, while deployment on a Unitree G1 verified zero-shot transfer across sparse footholds, wide gaps, and obstacle-rich environments. Current limitations include the map's local sensing range, reconstruction artifacts, and real-time mapping cost. Future work will investigate more efficient, temporally extended, and multi-scale perception.







\bibliographystyle{IEEEtran}
\bibliography{IEEEabrv, mybib}

\end{document}